# Designing, 3D Printing of a Quadruped Robot and Choice of Materials for Fabrication

**Akash Maity**
Metallurgical and Materials Engineering, National Institute of Technology Durgapur
Mahatma Gandhi Rd, A-Zone, Durgapur-713209, West Bengal, India
E-mail ID: am.20150292@btech.nitdgp.ac.in

**Koustav Roy**
Mechanical Engineering, National Institute of Technology Durgapur
Mahatma Gandhi Rd, A-Zone, Durgapur-713209, West Bengal, India
E-mail ID: kr.20150229@btech.nitdgp.ac.in

**Dhrubajyoti Gupta**
Mechanical Engineering, National Institute of Technology Durgapur
Mahatma Gandhi Rd, A-Zone, Durgapur-713209, West Bengal, India
E-mail ID:dg.17u10024@btech.nitdgp.ac.in

**Abstract-**

**Purpose-** This paper is based on design of a quadruped robot and manufacturing it with 3-D printer followed by its detailed analysis. It focuses on the advantages of additive manufacturing rather than conventional manufacturing techniques and also highlights its limitations. The consequences of choice of different materials for 3-D printing are evaluated in this report.

**Design/methodology/Approach-** The parts were designed in CAD software and made into Stereo lithography files, which were fed into the 3-D printing software. They were printed with the selection of ABS material. Low torque servos were employed in the beginning of the assembly and were controlled with the Arduino Uno microcontroller.

**Findings-** The versatility of legged robots require materials with high strength to weight ratio, where 3003 aluminium alloy sheets proved to be a better choice than conventional ABS. The ABS legs buckled under load and proved to be an inferior material choice for fabrication. The aluminium sheet fabricated legs not only imparted structural stability to the robot but also allowed in selection of more powerful servos for added strength. The bot could now achieve significant level of stability.

**Research Limitations/Implications-** The power supply used for powering the bot was of SMPS type of power supply which made it less mobile. Due to the huge demand of current by the high torque servos used later, the power supply became a huge limitation which can be overcome by applying Lithium ion batteries.

**Practical Implications-** The bot can be used for SLAM and autonomous navigation in areas where it is almost impossible for humans to access into.

**Originality/Value-** This paper shows a concrete study on efficacy of quick CAD designing and rapid fabrication.

**Keywords: 3-D Printing, Kinematics, EDM, Sheet Metal, Quadruped.**

## I. INTRODUCTION

Due to increased number of degrees of freedom, legged robots, in general, a far more complex than wheeled robots which are rather simple and easy to control. But wheeled robots have numerous backlogs in terms of their movement in undulating terrain, crossing hindrance, etc.[1], [5].Legged robots effectively provide solution to such problems. This paper is focused on a very popular legged robot, the quadruped robot. Solid Works CAD software is used to design the parts of the robot and then it is manufactured using FlashforgeCreatorPro3-D printer. This is the salient part of the paper as conventional manufacturing techniques were avoided which made possible in designing customized parts rather than relying on the standard available ones. It was found that the material choice was not optimal for supporting the load of the servos and the microcontroller after the bot was fully assembled. The Tibia (Fig. 1. b) began to buckle under load and got deformed, leading to stability issues. The Tibia was then redesigned and was re-fabricated with aluminium plates with the help of EDM, which resulted in higher strength to weight parts. Then the detailed analysis was carried out in terms of some standard parameters such as kinematics, stability, etc. The choice of material further allowed us to incorporate heavier, more powerful servos in order to impart more strength to the kinematics of the bot.

## II. LITERATURE SURVEY

The first idea of creation of legged robots was inspired from various species pre-existing in the environment such as centipedes, millipedes, etc. Subsequently some crude walking mechanism were formulated in order to accomplish the task. The first walking mechanism was developed by Chebyshev in 1870 which was a manifestation of four bar mechanism[4]. Subsequently, gaits were formulated and studied based on mathematical models in an attempt to improve the features of walking mechanisms. Thus, stability measurements and gait generation algorithms were created based on ideal cases, and improvements were eventually made leading to the present condition of such technology.

Our research is to create a simplified version of all the complex analysis and formulations that have been done in order to develop a quadruped robot. We have tried to simplify things in terms of design, manufacturing and analysis of this particular type of legged robot, the quadruped robot.

### III. PROBLEM DEFINITION

Modern technology is developed keeping in mind that it would assist or be of use to mankind. The prime problems which are faced by robot makers are cost, feasibility of manufacturing, resources in terms of material availability and many other things. In the recent years a significant development in the field of affordable 3-D printing has been witnessed. This has eased up the process of fabrication of 3D models which now can be produced at unprecedented frequencies. The advancements in 3D printing technology have motivated many people to get into CAD and CAM and design their own models. Those models now can be easily fabricated once the solid model has been created.

This report is aimed at investigating a similar topic as stated above. The following pages exhibit the process of fabrication of a quadruped robot. First, the parts are designed in CAD software. The parts are then converted into Stereo-Lithographic Image files which are then fed into a 3D printer software. This is followed by a brief discussion on material selection in 3D printing and numerous problems that can arise with it. Let us discuss about the kinematic analysis of the leg-joints and the joint trajectory planning.

### IV. METHODOLOGY

#### A. Hardware and software used

1. CAD software: CAD stands for Computer Aided Design. The name itself suggests that it is the use of computer systems to design, analyse and optimise a particular product. Some CAD softwares can even simulate the working of a particular system. Any open source or paid CAD software can be used out of the bundles available. The parts of the robot are designed in CAD software.

2. Flashforge Creator Pro 3D Printer: Thanks to the open source technology, this printer has provided precise and high quality 3D printed models. The CREATOR PRO is a Fused Deposition Moulding (FDM) type 3D printer. The sturdy metal frame seen in the product increases stability of the printer's moving parts. Metal platform support and 10mm Z-axis guide rods allow for precise movement on the Z-axis. The build plate is made from 6.3mm thickness alloy of aluminium.

It has 227x148x150 mm build volume. Layer resolution offered by this printer is 100-500 microns. The product weighs about 17 kgs. The integrated software it uses is ReplicatorG. A solid model file, after being created, can be transformed in Stereo-Lithographic Imaging File (STL) and fed into the 3D printer via USB or SD card. The build time will depend upon size, complexity and amount of material to be used in the model [6].

3. EDM: Electrical discharge machining, also known as spark machining, spark eroding, wire burning or wire erosion, is a manufacturing process whereby a desired shape is obtained by using electrical discharges. Material is removed from the work piece by a series of rapidly recurring current discharges between two electrodes, separated by a dielectric liquid and subjected to an electrical voltage. One of the electrodes is called the tool electrode, while the other is called the work piece electrode. When the voltage between the two electrodes is increased, the intensity of the electric field in the volume between the electrodes become greater than the strength of the dielectric, which breaks down, allowing current to flow between the two electrodes. As a result material gets removed, which gets carried away by draining liquid dielectric and fresh liquid dielectric is replaced[7].

#### B. Design and Fabrication of the quadruped robot

As stated earlier this is a solely self-designed robot. Therefore the first step was to design the parts that would constitute the robot. The robot to be designed would have 3 servo motors on each leg, therefore, giving it 3 degrees of freedom[1]. The legs would further have sub parts viz. 'upper leg' (Femur) and 'lower leg' (Tibia). These legs would then be attached to the base plate or main body. The following parts were designed in CAD software

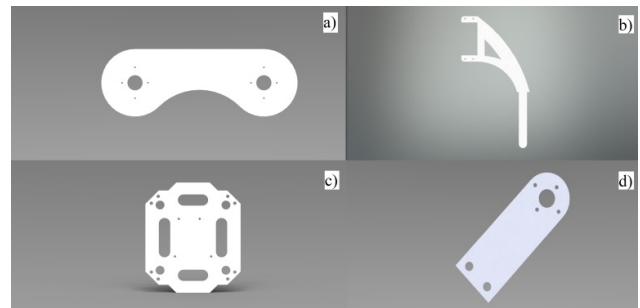

**Fig.1 Part by part 3-D model designed in CAD software a) Femur b) Tibia c) Body d) Extension Hand**

Due to build volume limitations in the 3D printer a separate 'Extension hand' part was designed which was to be screwed in on the four corners of the base plate. They were assembled in CAD software. The assembly is shown below-

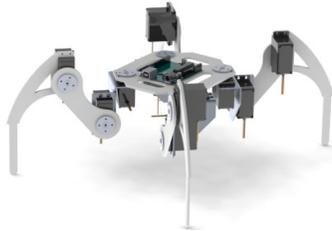

**Fig.2 Complete 3-D Constrained Assembly in CAD Software**

The 'Extension hand' part was later discarded as it lead to unnecessary instability issues. The solid models were converted into STL files and loaded into the 3-D printer. The robot had 4 legs. Therefore, as set of 4 upper and lower leg parts were fabricated and finally a base plate was fabricated. The 3D printer supported Poly Lactic Acid (PLA) and Acrylonitrile Butadiene Styrene (ABS) as materials for 3D printing. Though ABS had poor mechanical properties, it was the material of choice as PLA had poor machining properties and was resistant to adhesives. The 3-D printed parts are shown below

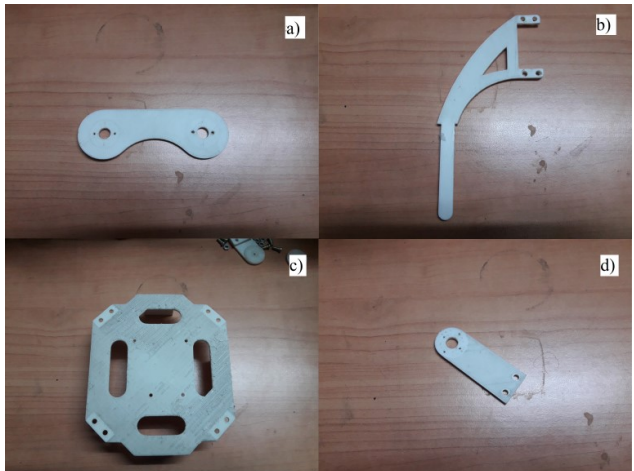

**Fig.3 Complete 3-D Printed Parts with ABS
a) Femur b) Tibia c) Body d) Extension Hand**

The holes were machined to satisfy the mating conditions of servo horns and screws. After each fabrication, the practice was to cut out the part from its raft, which formed the base support upon the build bed. Several parts got bent due to this as at that time they were still hot.

**A. Electronics and Micro-controllers**

Now let us focus on the control system of the robot. The robot has servo motors as its actuators, which are controlled by Arduino Mega. The Arduino Mega 2560 is a microcontroller board based on the ATmega2560. It has 54 digital input/output pins (of which 15 can be used as PWM outputs), 16 analog inputs, 4 UARTs (hardware serial ports), a 16 MHz crystal oscillator, a USB connection, a power jack, an ISCP header and a reset button.

Arduino Mega has been chosen in order to cop up with the requirement of number of digital I/O pins. In this context, it is worth mentioning that an effective and proper power supply has to be chosen for smooth and safe functioning of the servo motors as well as the Arduino board. In order to provide external power in an integrated manner, a compatible sensor shield was used. For initial purposes a computer grade power supply was chosen as it had the capability to supply high amperes of current at 12V. The servos that were to be installed in the bot were low torque HITEC HS 311 servos which demanded 6V and 1A at full load. So, an appropriate step down voltage regulator and power supply were used.

**B. Assembly of the robot**

The hands were first screwed in to the corners of the basal plate. Servos were attached to the legs and finally for the first time the bot was beginning to take shape.

The rough surface at the top was the result of the raft upon which the basal plate was fabricated. The software of the 3D printer automatically suggested the formation of raft for easy removal of the fabricated parts from the build plate.

The servos were assembled at the leg joints and each of the legs was attached to the servos situated at the four corners of the basal plate with the help of small aluminium bars which were machined for this purpose.

The servos were connected to the shield, which was in turn mounted on the Arduino Mega microcontroller.

After the assembly was complete, it was time to test the bot for its stability. In order to perform that, a simple code was uploaded to the microcontroller which held the servos at specific angles so that the bot could stand.

Upon performing initial tests, it was found that there were massive stability issues. The bot could stand but it would shake from time to time as the servos would fluctuate. This gave an insight that bot was barely able to lift its own weight! What went wrong?

**C. Material and Servo reselection**

There were two flaws evident in the process viz. selection of low torque 6W servo and 50% infill ABS as the structural material.

The servos were unable to generate enough torque to prevent mild slipping of the foot tip when placed on platform which led to vibrations and excessive heat generation on the servos, and hence, our actuators needed replacement.

This problem was not alone. The repeated and tedious vibration cycles created fatigue in the leg pieces: Tibia and Femur. The legs got twisted and due to unequal distribution of load bending stress was induced in the legs which started causing bending failure.

It was realised that there were 3 main problems with the robot:

Insufficient torque supplied by the servos

Thin design of the foot tip which resulted in less area of contact between the tip and the platform which in turn lead to increased chances of slipping

Material strength and durability

So, the team decided to go for re-fabrication of the leg-pieces using aluminium (which were manufactured with the help of EDM machine) thus ensuring better tensile strength and fatigue resistance and also replacing the low torque servos with high torque Futaba bls175sv with current specification of 1 A and voltage specification of 7.4V.

With respect to the good selection criterion the changes worked well, minimizing the repeated vibration and heating problem of the servos. But this was not the complete solution to our problem, the slipping of the foot-tip continued due to its thin design i.e. less contact area between the tip and the platform.

Now it was decided to use a simple glue-gun on the foot-tip to increase the area of contact and reduce the smoothness due to the good surface finish thus increasing the value of the friction coefficient. But this was also not the end of our troubles. Now we needed proper servo brackets to fit in the servo motors whose fitting holes were already failing due to shear. Thus, a simple cuboidal bracket was made to fit in the servos and now the design was stable enough to withstand its own load and stand for a span of about 12-13 minutes at stretch with minimal heating problem.

**D. Mathematical Modelling of the problem**

This section will put light on the detailed kinematic and dynamic analysis of the robot.

The main challenges are the planning and implementation of the legs' motions in generating a walking sequence, namely, how to plan the steps of the robot so that it moves in a desirable way while maintaining equilibrium constraints (quasi-static walking)[4]. It will be achieved by calculating the joint variables (simply speaking, angles turned by the servos) so as to achieve the desired motion of the tip of the leg. In this context it is quite evident that not only the individual legs' motions are to be coordinated but also there has to be a smooth coordination among the 4 legs themselves.

Kinematic Analysis: As stated earlier the robot has a base plate to which all the legs are attached which are similar and symmetric about the centre of mass axis of the base plate. Each leg has three servo powered rotary joints with the typical articulated (RRR) configuration, i.e. the second and third joints' axes are parallel to each other and perpendicular to the first joint's axis.

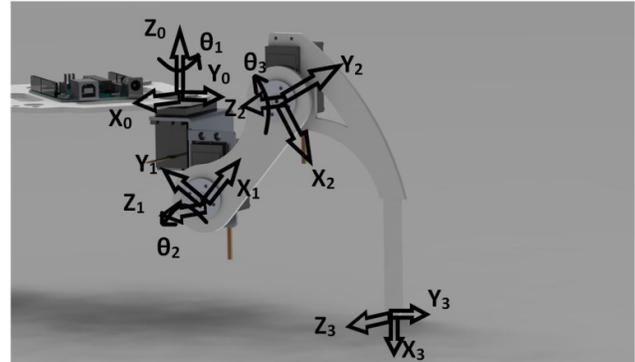

**Fig.4 Kinematic Diagram of a Leg of the Robot**

The kinematic analysis of any robot has got 2 phases viz. Forward kinematics and Inverse kinematics. The former deals with determination of the position of the end effector with known joint variables while the later deals with calculating the required joint variables in order to reach a particular position[3]. An important point is to be noted that inverse kinematics cannot be carried out without writing the forward kinematic equations. The detailed operations of all those in accordance with the robot are as follows:

The frames have been attached to each of the joints according to the DH parameter rules. The problem is governed by four variables, namely link length ($a_i$), link twist ($\alpha_i$), joint distance ($d_i$), and joint angle ($\theta_i$), required to completely describe the leg mechanism.

The general transformation matrix according to the DH parameter rule can be written as:

$$_{i-1}^{i}T = \begin{bmatrix} \cos(\theta_i) & -\sin(\theta_i)\cos(\alpha_i) & \sin(\theta_i)\sin(\alpha_i) & a_i\cos(\theta_i) \\ \sin(\theta_i) & \cos(\theta_i)\cos(\alpha_i) & -\sin(\alpha_i)\cos(\theta_i) & a_i\sin(\theta_i) \\ 0 & \sin(\alpha_i) & \cos(\alpha_i) & 0 \\ 0 & 0 & 0 & 1 \end{bmatrix}$$

In our case i will have values 1, 2, 3. By putting these values turn by turn we can find the individual transformation matrices. The final transformation matrix can be obtained by multiplying those matrices, i.e.

$$_{3}^{0}T = {_{1}^{0}T} \cdot {_{2}^{1}T} \cdot {_{3}^{2}T}$$

The final transformation matrix after multiplication of those transformation matrices is:

$$\begin{bmatrix} \cos(\theta_1)\cos(\theta_2+\theta_3) & -\cos(\theta_1)\sin(\theta_2+\theta_3) & -\sin(\theta_1) & (a_1+a_2\cos(\theta_2)+a_3\cos(\theta_2+\theta_3))\cos(\theta_1) \\ \sin(\theta_1)\cos(\theta_2+\theta_3) & -\sin(\theta_1)\sin(\theta_2+\theta_3) & -\cos(\theta_1) & (a_1+a_2\cos(\theta_2)+a_3\cos(\theta_2+\theta_3))\sin(\theta_1) \\ \sin(\theta_2+\theta_3) & \cos(\theta_2+\theta_3) & 0 & a_2\sin(\theta_2)+a_3\sin(\theta_2+\theta_3) \\ 0 & 0 & 0 & 1 \end{bmatrix}$$

The final coordinates of the foot tip can be expressed as:

$$(a_1 + a_2\cos(\theta_2) + a_3\cos(\theta_2 + \theta_3))\cos(\theta_1) = X$$
$$(a_1 + a_2\cos(\theta_2) + a_3\cos(\theta_2 + \theta_3))\sin(\theta_1) = Y$$
$$a_2\sin(\theta_2) + a_3\sin(\theta_2 + \theta_3) = Z$$

Where (X, Y, Z)' is the final position of the foot tip.

Similarly the values of $\theta_i$ can be calculated using the inverse kinematics. After simple algebraic manipulations the results are formulated below:

$$\theta_1 = atan2(Y, X)$$
$$\theta_2 = atan2(C, \sqrt{A^2 + B^2 + C^2}) - atan2(A, B)$$
$$\theta_3 = \frac{(\sqrt{X^2 + Y^2} - a_1)^2 + Z^2 + a_2^2 - a_3^2}{2a_2 a_3}$$

Where
$$A = 2a_3(X^2 + Y^2 - a_1)$$
$$B = 2Za_2$$
$$C = (X^2 + Y^2 - a_1)^2 + Z^2 + a_2^2 - a_3^2$$

Foot trajectory planning: After having determined, what must be the joint variables in order to move the robot to a particular position, it is the ardent need of the hour to ensure that the robot does not suffer any abrupt changes in its joint variables, which may lead to unsteady motions and fatal vibrations. Therefore, trajectory planning becomes necessary. It involves formulation of a sequence of movements that must be made to create a controlled movement between motion segments[2].

To ensure a smooth path to be followed by the swing foot, the joint trajectory is assumed to be a 3rd power polynomial function as the trajectory function of the joint angles $\theta_i$

$$\theta_i = c_{0i} + c_{1i}t + c_{2i}t^2 + c_{3i}t^3$$

Differentiating both sides,
$$\dot{\theta}_i = c_{1i} + 2c_{2i}t + 3c_{3i}t^2$$

Where i= 1, 2, 3
From velocity analysis,
$$J\dot{\theta} = \dot{P}$$
$$\dot{\theta} = J^{-1}\dot{P}$$

Where
$$P = \begin{bmatrix} X \\ Y \\ Z \end{bmatrix} \quad \theta = \begin{bmatrix} \theta_1 \\ \theta_2 \\ \theta_3 \end{bmatrix}$$
and
and J is the Jacobian matrix represented as

$$J = \begin{bmatrix} (a_1 + a_2\cos(\theta_2) + a_3\cos(\theta_2 + \theta_3))\sin(\theta_1) & -(a_2\sin(\theta_2) + a_3\sin(\theta_2 + \theta_3))\cos(\theta_1) & -(a_3\sin(\theta_2 + \theta_3))\cos(\theta_1) \\ (a_1 + a_2\cos(\theta_2) + a_3\cos(\theta_2 + \theta_3))\cos(\theta_1) & (a_2\sin(\theta_2) + a_3\sin(\theta_2 + \theta_3))\sin(\theta_1) & -(a_3\sin(\theta_2 + \theta_3))\sin(\theta_1) \\ 0 & a_2\cos(\theta_2) + a_3\cos(\theta_2 + \theta_3) & a_3\cos(\theta_2 + \theta_3) \end{bmatrix}$$

## V. RESULTS & DISCUSSIONS

We have used additive manufacturing instead of conventional manufacturing techniques, which obviously reduced the cost to some extent and increased the design innovation to a great extent. It is also less tedious compared to the conventional techniques. The robot faced some significant stability issues which were then resolved by replacing the servos and the ABS legs with aluminium legs. Some aluminium bars were also designed to connect between the body and the servo legs that further enhanced the stability of the robot. Glue (from the hot glue gun) was used on the foot tips to increase friction between the tips and the platform in order to prevent further slipping.

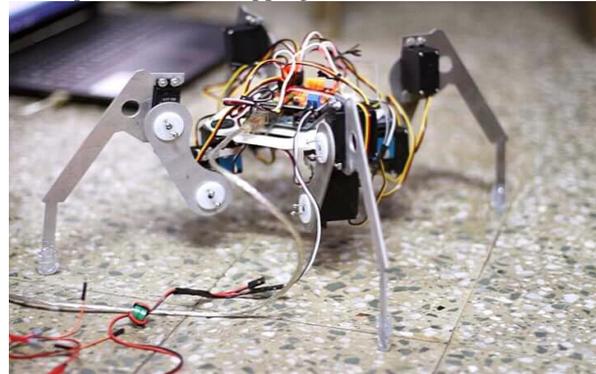

**Fig.5 Stable Standing Robot Made With Aluminium Legs**

There were some power supply issues too which were provided with some adequate temporary solutions.

## VI. CONCLUSION

So, our self-designed and self-manufactured robot worked quite well despite having some limitations. We are looking forward to use a lithium ion power supply instead of a switching mode power supply which would allow us to carry out the gaiting analysis in a more extensive way. A separate servo bracket has also been designed, that is to be manufactured with the help of aluminium sheet metal, which provide a further more stable structure to the robot.

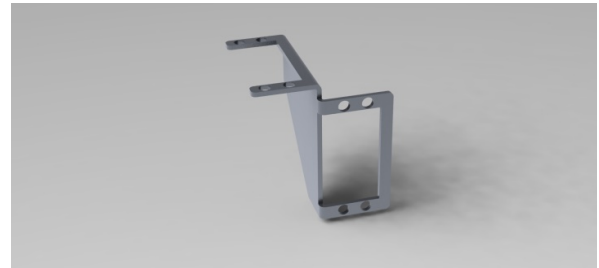

**Fig.6. Redesigned and Improved Servo Bracket**

## VII. FUTURE SCOPES

This paper can be significantly carried forward towards numerous manifestations. Some examples are: Correction of gaiting methods of the robot followed by enhanced locomotion like stair climbing with the help of SLAM (Simultaneous localization and Mapping) using LIDAR sensor, imitation of walking stance of a real spider and generalisation of energy efficient method for walking on uneven terrain. This particular robot's applicability can be well extended to human body detection at any natural hazard trodden site. Furthermore the robot can be embedded with numerous sensors to achieve some demand specific tasks.

**ACKNOWLEDGEMENT**

First of all, we would like to thank the authorities at CSIR-CMERI for their continued support, help and for providing the required materials and electronic components. Next the authors would like to thank the CAD-CAM lab, Department of Mechanical Engineering, National Institute of Technology Durgapur for letting us use the 3-D printer and providing us with the required plastic material ABS as and when required.

Next the authors would like to indebt their gratitude to Dr.AtanuMaity, HOD, Engineering Design Group, CSIR-CMERI and Dr.ShibenduSekhar Roy, Professor, NIT Durgapur for their continuous support and guidance regarding the decision making and engineering analysis.

Finally we would like to thank our friends, family and other fellow students for their continuous involvement and moral support towards making this project a success.